%% file: main.tex
\pdfoutput=1

\documentclass[11pt]{article}

\usepackage[final]{coling}

\usepackage{times}
\usepackage{latexsym}

\usepackage[T1]{fontenc}

\usepackage[utf8]{inputenc}

\usepackage{microtype}

\usepackage{inconsolata}

\usepackage{graphicx}

\usepackage{CJKutf8}    
\usepackage{pifont}        
\usepackage{booktabs}  
\usepackage{multirow}   
\usepackage{stfloats}     
\usepackage{amsmath}  
\usepackage{amsfonts}  
\usepackage{xcolor}        
\usepackage[normalem]{ulem} 
\newcommand\hl{\bgroup\markoverwith
  {\textcolor{grey!15}{\rule[-.5ex]{2pt}{2.5ex}}}\ULon}
\usepackage[most]{tcolorbox}

\title{Chain-of-Tools: Utilizing Massive Unseen Tools in the CoT Reasoning of Frozen Language Models}

\author{Mengsong Wu \hspace{1cm} Tong Zhu \hspace{1cm} Han Han \hspace{1cm} {\bf Xiang Zhang} \\
{\bf Wenbiao Shao} \hspace{1cm} {\bf Wenliang Chen} \\
Soochow University, Shizi Street 1, 215006 Suzhou, China \\
  \texttt{\{mswumsw,tzhu7,hhan,xzhangxzhang23,wbshao\}@stu.suda.edu.cn, wlchen@suda.edu.cn} \\}

\begin{document}
\begin{CJK}{UTF8}{gkai}

\maketitle

\input{sections/abstract}

\input{sections/introduction}

\input{sections/relatedwork}

\input{sections/methodology}

\input{sections/experiment}

\input{sections/conclusion}

\input{sections/limitations}

\bibliography{custom.bib}

\input{sections/appendix}

\end{CJK}
\end{document}

%% file: sections/abstract.tex
\begin{abstract}
\label{section:abstract}

Tool learning can further broaden the usage scenarios of large language models (LLMs).
However most of the existing methods either need to finetune that the model can only use tools seen in the training data, or add tool demonstrations into the prompt with lower efficiency.
In this paper, we present a new Tool Learning method \textbf{Chain-of-Tools}.
It makes full use of the powerful semantic representation capability of frozen LLMs to finish tool calling in CoT reasoning with a huge and flexible tool pool which may contain unseen tools.
Especially, to validate the effectiveness of our approach in the massive unseen tool scenario, we construct a new dataset SimpleToolQuestions.
We conduct experiments on two numerical reasoning benchmarks (GSM8K-XL and FuncQA) and two knowledge-based question answering benchmarks (KAMEL and SimpleToolQuestions).
Experimental results show that our approach performs better than the baseline.
We also identify dimensions of the model output that are critical in tool selection, enhancing the model interpretability.
Our code and data are available at: \url{https://github.com/fairyshine/Chain-of-Tools} .

\end{abstract}

%% file: sections/introduction.tex
\section{Introduction}
\label{section:introduction}


The development of autonomous agent systems (\citealp{agent_survey_gaoling, agent_survey_fudan}), propelled by real-world applications (\citealp{GPT4}) of Large Language Models (LLMs), has become a popular focus in both academic and industry communities.
Benefit from LLM's emergent ability (\citealp{emergent_ability, emergent_LLM_survey}) to think questions comprehensively and integratedly, LLM agent may give brilliant step-by-step solutions during multiple-turn chat with users.
Despite LLM is expert in logical reasoning and breaking down problems, it can't accomplish a lot of specific tasks like calculating math formulas or drawing paintings.
In order to extend the application scenarios, equipping LLM agent with external tools is a reasonable solution.
That's what the Tool Learning (\citealp{tool_learing_with_foundation_model}) task investigates: how to make LLMs better utilize tools in the process of reasoning?

\input{figures/agent\_framework}

There are two kinds of typical methods for Tool Learning.
(1) Fine-tuning based methods like API-Bank (\citealp{li-etal-2023-api}) and ToolLLM (\citealp{ToolLLM}) can efficiently and precisely call tools which have been seen during training, while the general capabilities of LLMs such as emergent ability and Chain of Thought (CoT) might be influenced by fine-tuning (\citealp{CoT}).
ToolkenGPT (\citealp{ToolkenGPT}) introduces a method which only fine-tunes extra tool-token embeddings without hurting the original model, but it still can't use unseen tools.
(2) In-context learning (ICL) based methods like HuggingGPT (\citealp{HuggingGPT}) and AgentBench (\citealp{AgentBench}) are flexible to call unseen tools with ICL prompt, while it is less efficient in reasoning when given massive tools. 
The above methods in Table~\ref{table:paradigm_comparison} have shown a certain success in utilizing tools in the LLM agents.
However, we argue that the LLM agent should be capable of efficiently managing a large amount of tools and fully utilizing unseen ones during the CoT reasoning, as many new tools may emerge daily in real-world application scenarios.

\input{tables/paradigm_comparison}

In this paper, we introduce \textbf{C}hain-\textbf{o}f-\textbf{Tools} (\textbf{CoTools}), a brand new fine-tuning based Tool Learning method.
We follow the way of fine-tuning based method since it is much more efficient to call tools which is critical for practical applications.
The remaining problem is how to effectively utilize unseen tools in the process of CoT reasoning without hurting the model's capability.
In order to address this problem, we design the ideal Tool Learning procedure shown in Figure~\ref{figure:agent_framework}.
CoTools fully utilizes the token semantic representation, which is often called as the hidden states generated by LLMs, as input to judge where to call tools and select which tools to call (e.g. "Weather" tool in Figure~\ref{figure:agent_framework}) then the tool calling result is added in the answer (e.g. "sunny" in Figure~\ref{figure:agent_framework}).
First, the user query is prompted with ICL and CoT before being inputted into the LLM.
Then CoTools judges whether to call a tool with the hidden state of new answer token when the LLM is generating every answer token.
If tool calling is needed, CoTools calculates query vectors and tool vectors respectively with corresponding hidden states given for tool selection.
The tool vectors of unseen tools can also be computed from their description for flexible retrieval.
What's more, since the LLM is frozen, its CoT reasoning ability remains unaffected.

Contributions of this paper are listed as following:
\begin{itemize}
\item The new Tool Learning method \textbf{CoTools} can utilize massive unseen tools efficiently in the process of CoT reasoning with the frozen LLM. 
It fully explores the generation capability and the semantic representation capability of the LLM for better Tool Learning procedure.
Unseen tools can be easily equipped for tool selection with their detailed descriptions without introducing external retriever.
\item We construct the dataset \textbf{SimpleToolQuestions} (STQuestions) including 1836 tools to evaluate the tool selection performance of each method.
Compared with former benchmarks, it focuses on evaluating in the massive unseen tool scenario.
\item For detailed evaluation, we conduct experiments on four benchmarks: GSM8K-XL, FuncQA, KAMEL and STQuestions. 
Experimental results show that CoTools performs better than baseline in both numerical reasoning and knowledge-based question answering.
Moreover, we discover the key dimensions of the hidden states for tool selection, which may enhance the interpretability of the model.
\end{itemize}

%% file: figures/agent_framework.tex
\begin{figure}[htb]    
	\centering   
	\includegraphics[width=0.98\linewidth]{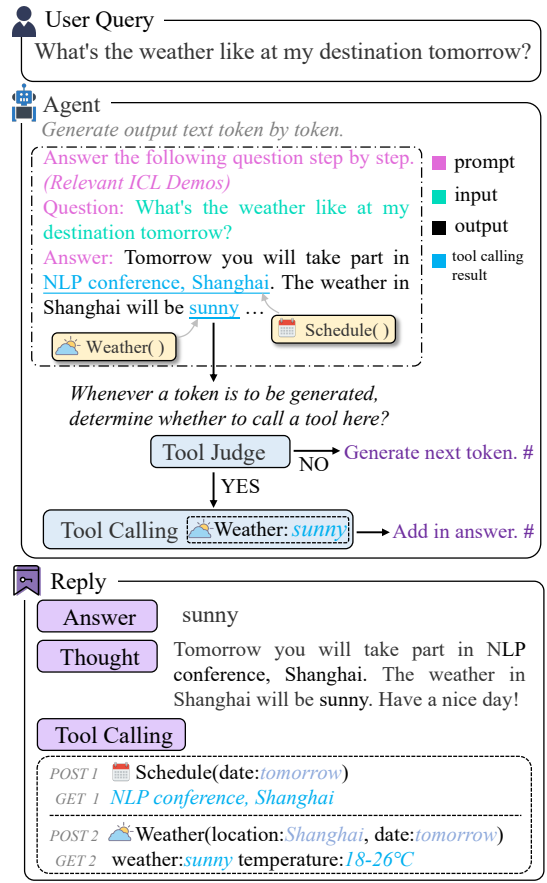}    
	\caption{The ideal tool calling procedure.Take the input query "What's the weather like at my destination tomorrow?" as an example.} 
	\label{figure:agent_framework}
\end{figure}

%% file: tables/paradigm_comparison.tex
\begin{table}[htb]
\centering
\resizebox{0.48\textwidth}{!}{
\begin{tabular}{lccccc}
\toprule
\textbf{\begin{tabular}[c]{@{}c@{}}Tool Learning \\ Paradigms\end{tabular}} & \textbf{\begin{tabular}[c]{@{}c@{}}Frozen \\ LMs\end{tabular}} & \textbf{Plugable} & \textbf{\begin{tabular}[c]{@{}c@{}}Massive \\ Tools\end{tabular}} & \textbf{\begin{tabular}[c]{@{}c@{}}Unseen \\ Tools\end{tabular}} & \textbf{\begin{tabular}[c]{@{}c@{}}Ability to \\ Use Extensive Data\end{tabular}} \\ \midrule
\textbf{Fine-tuning} & {\color{red}\ding{56}} & {\color{red}\ding{56}} & {\color{red}\ding{56}} & {\color{red}\ding{56}} & {\color[HTML]{008114}\ding{52}} \\
\textbf{In-Context Learning} & {\color[HTML]{008114}\ding{52}} & {\color[HTML]{008114}\ding{52}} & {\color{red}\ding{56}} & {\color[HTML]{008114}\ding{52}}  & {\color{red}\ding{56}} \\
\textbf{ToolkenGPT} & {\color[HTML]{008114}\ding{52}} & {\color[HTML]{008114}\ding{52}} & {\color[HTML]{008114}\ding{52}} & {\color{red}\ding{56}} & {\color[HTML]{008114}\ding{52}} \\
\textbf{CoTools(Ours)} & {\color[HTML]{008114}\ding{52}} & {\color[HTML]{008114}\ding{52}} & {\color[HTML]{008114}\ding{52}} & {\color[HTML]{008114}\ding{52}} & {\color[HTML]{008114}\ding{52}} \\
\bottomrule
\end{tabular}
}
\caption{Comparision of the mainstream Tool Learning paradigms. "Plugable" means that tools can be flexibly loaded. "Ability to use extensive data" means that training data can be used to improve the performance of the method. (The table is partially referenced from Table 1 in the paper of ToolkenGPT.)}
\label{table:paradigm_comparison}
\end{table}

%% file: sections/relatedwork.tex
\section{Related Work}
\label{section:relatedwork}

\subsection{Tool Learning}

Tool Learning (\citealp{tool_learing_with_foundation_model}) enables foundation models to leverage specialized external tools.
LLMs then can accomplish much more complex tasks in realistic scenarios.

\noindent\textbf{Fine-Tuning based Tool Learning}

Most of these researches construct relevant tool datasets to fine-tune LLMs.
Toolformer (\citealp{Toolformer}) explores how to generate tool learning data with raw dataset given.
API-Bank (\citealp{li-etal-2023-api}) builds the comprehensive tool learning benchmark with tools in many fields.
GeneGPT (\citealp{GeneGPT}) uses NCBI web APIs as tools to validate the model's ability to call tools under a specific field.
Gorilla (\citealp{Gorilla}) takes online models from platforms like HuggingFace as tools to extend the usage scenarios of tool calls. 
ToolLLM (\citealp{ToolLLM}) generates dataset ToolBench with massive real world tools and also propose a multi-step tool learning inference method DFSDT.
API-BLEND (\citealp{API-BLEND}) aggregates the various types of datasets available with elaborate analysis.
ToolkenGPT (\citealp{ToolkenGPT}) doesn't adapt model weights but add many special tokens in the vocabulary.
One special token represents one external tool.
The corresponding tool is called when the model generates the special token.
ToolkenGPT just need to train the token embedding which is efficient.
CoTools does not train the foundation model but post-processing modules for the hidden states output by the model.

\noindent\textbf{In-Context Learning based Tool Learning}

In-context learning (\citealp{GPT3}) is one of the most prominent capabilities of the LLMs.
With the help of ICL, the model performance in the few-shot scenario of many tasks has improved dramatically.
ICL has become a basic trick in the wide usage of  LLMs.
There is no exception in the Tool Learning task.
TaskMatrix.AI (\citealp{Taskmatrix_ai}) proposes the multimodal conversational system to equip the foundation model with rich cross-modal tools from the API platform.
Huggingface models are also treated as tools in the paper of HuggingGPT (\citealp{HuggingGPT}). 
ToolDoc (\citealp{ToolDoc}) provides tool documentions as an alernative to tool demonstraions which are added in ICL prompt.
AgentBench (\citealp{AgentBench}) is a comprehensive agent benchmark for both commercial and open-sourced LLMs.
ToolTalk (\citealp{ToolTalk}) evaluates GPT-3.5 and GPT-4 (\citealp{GPT4}) with OpenAI API, which emphasizes tools that affect the external world such as sending emails.
TaskWeaver (\citealp{Taskweaver}) asks the model to call tools in the format of generating python code. Tools in it are python functions which can be executed.
ChatCoT (\citealp{chen-etal-2023-chatcot}) generates iterative tool-augmented reasoning according to the given problem. 
CoTools does not use the ICL prompt for tool selection, thereby enhancing efficiency in massive tool scenarios.
It uses the ICL prompt to fill in tool parameters with tool calling demonstrations given.

\subsection{Chain of Thought}

Prompt learning is developing rapidly with the rise of LLMs.
Chain of Thought is firstly proposed in \citet{CoT}. 
It encourages the model to generate thought about the question that leads to a better answer.
Zero-shot-CoT (\citealp{Zero-shot_CoT}) finds it work that add the phrase "Let's think step by step." in prompt. The phrase is now widespreadly used in the prompt engineering of LLMs.
There are also attempts to combine Zero-shot-CoT and Tool Learning together like ChatCoT and ToolkenGPT.
It's helpful to improve tool selection and invocation by models for complex problems.
Subsequent attempts can be made to utilize improved versions of CoT such as Tree of Thoughts (\citealp{ToT}) to complete Tool Learning task.

%% file: sections/methodology.tex
\section{Methodology}

In this section, we introduce Chain-of-Tools, a novel fine-tuning based Tool Learning method.
The core idea of CoTools is to leverage the semantic representation capabilities of frozen foundation models for determining where to call tools and which tools to call.
The foundation model $\mathcal{M}$ is usually an auto-regressive language model that would generate output token-by-token.
Given the input token list of length $n$ $[x_{1},x_{2},\dots,x_{n}]$ ($x \in \mathcal{V}$, $\mathcal{V}$ is the vocabulary of $\mathcal{M}$) which is tokenized from the input text, the model $\mathcal{M}$ can generate the hidden state of the last token, which is called $h_{n}$ ($h_{n} \in \mathbb{R}^{d}$, $d$ is the dim of the hidden states):
\begin{equation}
  \label{equation:foundation_model}
  h_{n} = \mathcal{M}(x_{n} ; x_{1},\dots,x_{n-1})
\end{equation}
Then we can get the next token $x_{n+1}$ caculated by the language model head $\mathcal{H}_{LM}$ of $\mathcal{M}$:
\begin{equation}
  \label{equation:language_model_head}
  x_{n+1} = \mathcal{H}_{LM}(h_{n})
\end{equation}
Instead of just being used to generate the next token as usual, the hidden state $h_n$ plays an important role in our method.

The main structure of CoTools contains 3 parts: Tool Judge, Tool Retriever and Tool Calling like in Figure~\ref{figure:method}.
The \textbf{Tool Judge} determines whether or not to invoke a tool during the CoT reasoning process.
The \textbf{Tool Retriever} selects the most suitable tool based on the query and the answer fragment that has been generated.
The \textbf{Tool Calling} fills in parameters of the selected tool with ICL prompt, executes it and gets the return value.

\input{figures/TAT\_method}

\input{figures/model\_architecture}

\subsection{Tool Judge: Whether Calling Tools}
\label{section:method_judge}

The Tool Judge $\mathcal{J}$ is used for determining whether to call a tool at the specific position of answer.
It is calculated with the input hidden states $h$:
\begin{equation}
  \label{equation:judge}
  \mathcal{J}(h)=W_{down}^{\mathcal{J}} ((\sigma W_{gate}^{\mathcal{J}}(h) )\otimes W_{up}^{\mathcal{J}}(h) ) 
\end{equation}
where $W_{gate}^{\mathcal{J}}, W_{up}^{\mathcal{J}}\in \mathbb{R}^{d \times D}$, $W_{down}^{\mathcal{J}}\in \mathbb{R}^{D \times 1}$ are parameters to be optimized. ($D$ is the hyper-parameter for the intermediate size).
$\sigma$ represents the activation function and $\otimes$ represents multiplication of corresponding positions.

The initial input token list is tokenized from the query with ICL and CoT prompt.
CoTools judges whether to call a tool whenever a new answer token is to be generated.
Suppose the current token list is $[x_{1},x_{2},\dots,x_{t}]$. We calculate the hidden state $h_{t}$ of the last token $x_{t}$ with Equation~\ref{equation:foundation_model}.
It is used to calculate the $Score_{\mathcal{J}}$:
\begin{equation}
  \label{equation:judge_2}
  Score_{\mathcal{J}} = \mathcal{J}(h_{t}) \in [0,1]
\end{equation}
If $Score_{\mathcal{J}}$ is larger than threshold $\theta$ (typically set to $0.5$), we attempt to call a tool here.
Otherwise, the model outputs the next token $x_{t+1}$ with Equation~\ref{equation:language_model_head}.

We train $\mathcal{J}$  as the sequence labeling task.
The objective function is the binary cross entropy loss:
\begin{equation}
  \label{equation:judge_3}
  \mathfrak{L}_{Judge} = \mathfrak{L}_{BCE} (Score_{\mathcal{J}},Label)
\end{equation}
where $Label \in \{0,1\}$.

\subsection{Tool Retriever: Find Needed Tools}
\label{section:method_retrieve}

The Tool Retriever consists of the Query Encoder $\mathcal{E_{Q}}$ and the Tool Encoder $\mathcal{E_{T}}$.
Both of them are used to calculate the vectors for retrieval.
The Query Encoder $\mathcal{E_{Q}}$ takes the hidden state of a token as input.
In order to keep as much information as possible in the original hidden states, we use residual connection which is really important for the final result.
$\mathcal{E_{Q}}'$ can thus be seen as an offset optimization of the original hidden states used for retrieval:
\begin{equation}
  \label{equation:retriever}
  \begin{aligned}
  \mathcal{E_{Q}}'(h) &= W_{down}^{\mathcal{Q}}((\sigma  W_{gate}^{\mathcal{Q}}(h)) \otimes W_{up}^{\mathcal{Q}}(h)) \\
  \mathcal{E_{Q}}(h) &= norm(W_{dim} \otimes (h + \mathcal{E_{Q}}'(h)))
  \end{aligned}
\end{equation}
where $W_{gate}^{\mathcal{Q}}, W_{up}^{\mathcal{Q}}\in \mathbb{R}^{d \times D}$, $W_{down}^{\mathcal{Q}}\in \mathbb{R}^{D \times d}$ and $W_{dim}\in \mathbb{R}^{d}$ are parameters to be optimized.
$\otimes$ represents multiplication of corresponding positions.
$norm$ represents tensor normalization.

When CoTools decides to call a tool in the process of reasoning, it tokenizes the query and answer fragment in retrieval prompt ending with the special END token $x_{end}$ (e.g. "</s>" in LLaMA2) to get $[x_{1},x_{2},\dots,x_{L_{q}},x_{end}]$.
The model $\mathcal{M}$ then calculates the hidden states of $x_{end}$ which is used for calculating the query vector $V_{Q}$:
\begin{equation}
  \label{equation:retriever_2}
	V_{Q} = \mathcal{E_{Q}}(\mathcal{M}(x_{end};x_{1},\dots,x_{L_{q}}))
\end{equation}

Suppose the external tool pool is $\mathbb{T}$.
For any tool $T \in \mathbb{T}$, we compute the tool vector $V_{T}$ in a similar way as we compute the query vector $V_{Q}$:
\begin{equation}
  \label{equation:retriever_3}
  \begin{aligned}
  \mathcal{E_{T}}'(h) &= W_{down}^{\mathcal{T}}((\sigma  W_{gate}^{\mathcal{T}}(h)) \otimes W_{up}^{\mathcal{T}}(h)) \\
  \mathcal{E_{T}}(h) &= norm(W_{dim} \otimes (h + \mathcal{E_{T}}'(h))) \\
  V_{T} &= \mathcal{E_{T}}(\mathcal{M}(x_{end};x_{1},\dots,x_{L_{t}}))
  \end{aligned}
\end{equation}
where $W_{gate}^{\mathcal{T}}, W_{up}^{\mathcal{T}}\in \mathbb{R}^{d \times D}$, $W_{down}^{\mathcal{T}}\in \mathbb{R}^{D \times d}$ are parameters to be optimized.
$\mathcal{E_{Q}}$ and $\mathcal{E_{T}}$ share the same parameter $W_{dim}$ since $W_{dim}$ is used to identify which dimensions of the hidden state play a role in the process of tool retrieval.

We dot multiply the query vecotor $V_{Q}$ and the tool vector $V_{T}$ to calculate the similarity score for the corresponding tool $T$.
The tool $T^{*}_{Q}$ with the highest score is what we needed here.
\begin{equation}
  \label{equation:retriever_4}
  Score_{Q,T} = V_{Q} \cdot V_{T}
\end{equation}
\begin{equation}
  \label{equation:retriever_5}
  T^{*}_{Q} = \underset{T}{argmax} \: Score_{Q,T}
\end{equation}

The Tool Retriever is trained with the constrastive learning method in batch as other retrievers do (\citealp{karpukhin-etal-2020-dense}).
The tool search during training is limited to the tools involved in the single data batch, rather than the entire tool pool during evaluation.
The objective function is the cross entropy loss:
\begin{equation}
  \label{equation:retriever_6}
  \mathfrak{L}_{Retriever} = \mathfrak{L}_{CE} (Score_{Q,T_{Batch}},Label)
\end{equation}

\subsection{Tool Calling: Use Retrieved Tools}
\label{section:method_call}

After finding the needed tool, the foundation model generates the parameters of it with ICL prompting.
The tool calling format should be emphasized in the prompt so that we can extract parameters value properly with regex expression from the model output.
Then the tool is executed and the result is added in answer.

%% file: figures/TAT_method.tex
\begin{figure*}[]
	\centering   
	\includegraphics[width=0.98\linewidth]{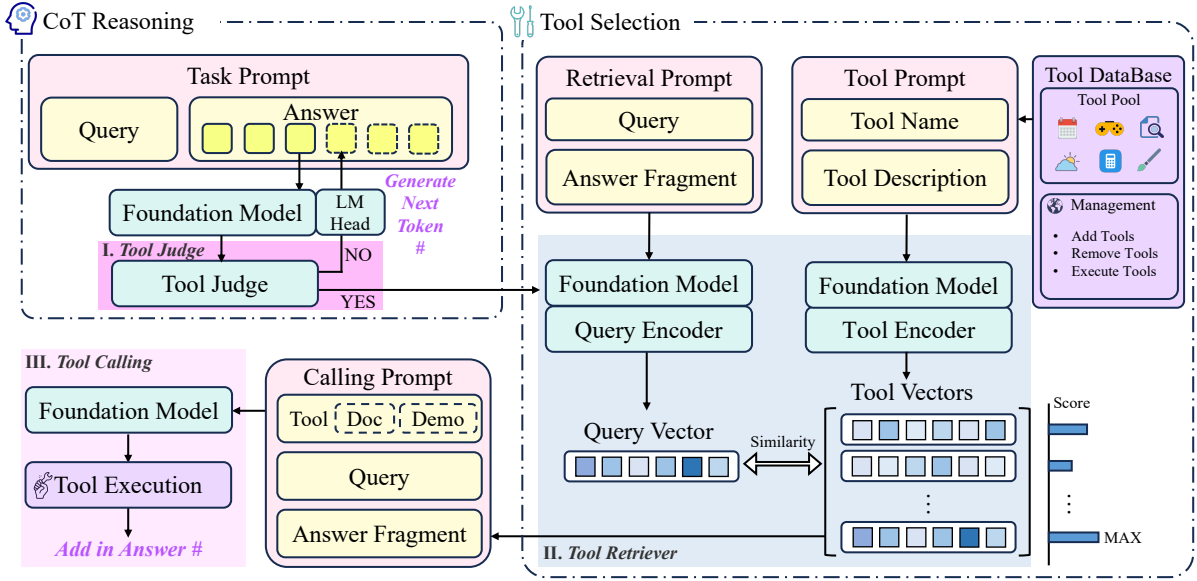}    
	\caption{Overview of the method CoTools. Just like the example in Figure~\ref{figure:agent_framework}, CoTools judges whether to call a tool whenever a new answer token is to be generated. The answer Fragment means the answer text that has been generated by the foundation model.}
	\label{figure:method}
\end{figure*}

%% file: figures/model_architecture.tex
\begin{figure}[]    
	\centering   
	\includegraphics[width=0.96\linewidth]{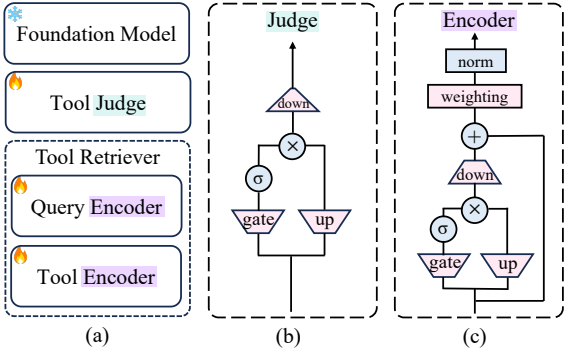}    
	\caption{Composition and structure of the model. The parameters of the foundation model are frozen. We only train the tool judge, the query encoder and the tool encoder.} 
	\label{figure:net}
\end{figure}

%% file: sections/experiment.tex
\section{Experiment}
\label{section:experiment}

In this section, we apply CoTools to two distinct Tool Learning application scenarios in English: calculating arithmetic questions and finding relevant knowledge in the Knowledge Base.
For the former, we use the numerical reasoning datasets GSM8K-XL and FuncQA which are created in ToolkenGPT (\citealp{ToolkenGPT}).
For the later, we select the Knowledge-Based Question Answering (KBQA) datasets KAMEL (\citealp{KAMEL}) and SimpleQuestionsv2 (\citealp{SimpleQuestions}).
To make the SimpleQuestionsv2 better suited for evaluation, we rewrite it using ChatGPT then get the new dataset SimpleToolQuestions (STQuestions).
Number of tools in these datasets is shown in Table~\ref{table:dataset_tool_amount}.

\input{tables/dataset\_tool\_num}

Experiments on these benchmarks show that CoTools outperforms baseline across the board.
Our method allows for more accurate tool selection even when the tool scale is very large.
Also it demonstrates generalization performance on unseen tools.
Besides, we find key dimensions of using hidden states for vector retrieval that help enhance the interpretability of the model.

\subsection{Experiment Setting}
\label{section:experiment_setting}

\subsubsection{Numerical Reasoning}
\label{section:experiment_setting_numerical_reasoning}

To evaluate tool learning paradigms on the numerical reasoning task, we use the two datasets created by ToolkenGPT (\citealp{ToolkenGPT}): 
(1) \textbf{GSM8K-XL}: It's an enhanced version of the existing GSM8K (\citealp{GSM8K}) dataset with 4 basic arithmetic operations ($+-\times\div$) as tools.
(2) \textbf{FuncQA}: It's an synthetic dataset with 13 arithemetic tools, with both one-hop and multi-hops questions.

Because the same arithmetic problem can often be solved in many different procedures, we focus on the correctness of the final result rather than on the specific tool calling process in the evaluation.
The evaluation metric is just like in ToolkenGPT.
We use the \textbf{Round Accuracy} metric for GSM8K-XL and FuncQA One-Hop test sets.
The float numbers are rounded to two decimals.
The \textbf{Approx Accuracy} metric is used for the FuncQA Multi-Hop test set.
It allows for errors of 0.1\% accuracy in multi-step calculations.

\subsubsection{Knowledge-Based Question Answering}
\label{section:experiment_setting_KBQA}

For evaluating whether tool learning paradigms can choose proper relevant tools  on the KBQA task, we use an existing dataset KAMEL and construct a new dataset SimpleToolQuestions with a much larger amount of tools.

(1) \textbf{KAMEL} (\citealp{KAMEL}): It's an QA dataset built with the knowledge in KB Wikidata. 
It has 234 relations which are viewed as tools.
We use the post-processing version of the paper ToolkenGPT (\citealp{ToolkenGPT}) with a gold training set KAMEL(\emph{sup}) and a synthetic training set KAMEL(\emph{syn}) generated by ChatGPT.

(2) \textbf{SimpleToolQuestions} (\textbf{STQuestions}): we construct it based on the KBQA dataset SimpleQuestionsv2 (\citealp{SimpleQuestions}).
One case in the raw dataset contains a question and a triplet (head entity, relationship, tail entity).
The raw data needs models to find the head entity in the question.
Then the knowledge subgraph about the head entity is founded in the KB Freebase.
Relationship is searched in a small scope.
But if we add all relations into the tool pool, it's not proper to retrieve tools in the whole with short questions.
There are many similar tools thus it is impossible to judge.
In order to make this dataset more applicable for tool learning evaluation, we rewrite the original question using ChatGPT.
The new questions provide a more detailed description of the needed relationship.
The dataset has 1836 tools with description from many different domains.
999 of them appear in the training set and 837 unseen tools are only in the test set.
We also generates tool description for each tool like the format of KAMEL.
It can be used to assess the accuracy of selection against seen and unseen tools in the large tool scale scenario, and whether the model favors tools of certain domains.

For KBQA benchmarks, we are more concerned about the accuracy of tool selection.
We want to evaluate whether the model is able to understand the semantic information of the questions and select the relevant tools, espcially when the tool pool is pretty large or the tools are unseen.

\subsection{Main Result}
\label{section:experiment_main_result}

Here is the main evaluation section of the paper.
We compare our method CoTools with baselines like 0-shot ChatGPT and ToolkenGPT.

\input{tables/main_result_math}

For the numerical reasoning task, the performances of different methods on the GSM8K-XL and FuncQA datasets are shown in Table~\ref{table:main_result_math}.
On the GSM8K-XL dataset, CoTools and ToolkenGPT based on LLaMA2-7B are comparable with ChatGPT in effect.
On the FuncQA dataset, CoTools is a little bit better than ToolkenGPT in one-hop questions with the foundation model LLaMA2-7B while makes even in multi-hops questions.
The performance of CoTools is very dependent on the foundation model's own ability.
It enhances the performance of foundation models across all ability levels compared to baseline methods, making the strong even stronger.

\input{tables/main_result_KBQA}

For KBQA benchmarks, the tool selection results of different methods are listed in Table~\ref{table:main_result_KBQA}.
For cases with large amounts of gold training data such as KAMEL(\emph{sup}), both methods perform well.
With synthetic training sets KAMEL(\emph{syn}) created by ChatGPT, CoTools does much better though it is still a long way from being usable.
CoTools does better in low-quality training data scenarios, perhaps because of the contrastive learning training approach.
Similarly CoTools performs better in massive tool (999 seen tools in the STQuestions dataset) scenarios.
For 837 unseen tools in the STQuestions dataset, CoTools can use the detailed description of tools for better retrieval while ToolkenGPT only can use tools that have been trained.
More details about unseen tools are further investigated in Section~\ref{section:experiement_unseen_tools}.

\subsection{Analysis}

\subsubsection{Data Synthesis}

The quality of dataset used for fine-tuning is very important (\citealp{LIMA}), so we want to investigate the effect of dataset generated by human and the LLM.
We use two versions of KAMEL training set for comparative validation.
(1) \textbf{KAMEL(\emph{sup})} is sampled from raw KAMEL training set.
It simulates the real-world scenario with sufficient in-domain training data.
(2) \textbf{KAMEL(\emph{syn})} is synthesized by ChatGPT.
For the situation where massive new tools are added to the tool pool, it is costly for human annotators to annotate sufficient data.
It is common today to use LLMs to assist in generating data for model tuning.

\input{figures/data\_synthesis\_sup}

With KAMEL(\emph{sup}), both methods perform pretty well in Figure~\ref{figure:data_synthesis_sup}.
This demonstrates that the tool selection subtask can already be well solved by existing methods when high quality training data are sufficient.
CoTools remains nearly 100\% correct with the TOP 5 tools selected so it would land well in real-world scenarios.

\input{figures/data\_synthesis\_syn}

For the synthetic training set KAMEL(\emph{syn}), Figure~\ref{figure:data_synthesis_syn} shows that CoTools is stronger than ToolkenGPT by more than 20\%.
Since the foundation model is frozen and the training data is limited, contrastive learning methods used in CoTools might be particularly effective. 
It can better teach the agent to distinguish between tools, even though the data quality is poor.

\subsubsection{The number of tools}

In this part, we try to explore the upper limit of the amount of loading tools supported by each method.
The benchmark STQuestions contains 999 tools which appear in the training set.
The evaluation results on it are shown in Figure~\ref{figure:scaling_law} below.

\input{figures/scaling\_law}

As the number of tools reaches a thousand scale, CoTools still outperforms the baseline by more than 10\%.
With the number of tools increases, the probability of similar tools appearing is much higher.
Benefit from the strong language understandng capabilty of LLMs, the tool learning method CoTools which better uses the LLM's own ability does better in this situation.

\subsubsection{Unseen Tools}
\label{section:experiement_unseen_tools}

It is meaningful to evaluate the generalizability of the fine-tuned model especially in tool learning.
It's inconvenient to adapt the model weight whenever some new tools are added.
As mentioned in Section~\ref{section:experiment_setting_KBQA}, total 837 tools in the dataset STQuestions do not appear in the training set.
We use these unseen tools as out-of-domain distributions to examine the generalization ability of the model.

\input{figures/unseen\_error\_analysis}

The main results on unseen tools have been listed in Table~\ref{table:main_result_KBQA}.
CoTools has a top1 accuracy of 10.41\% and a top5 accuracy of 33.68\%.
ToolkenGPT has a top1 accuracy of 0.0\%.
We count the wrong cases in the results as in Figure~\ref{figure:unseen_error_analysis}.
It's easy to see from the figure that ToolkenGPT has a clear preference for a small group of tools however CoTools does not.
In other words, CoTools focuses more on how to distinguish and identify tools during training instead of remembering them.

\subsubsection{Key Dimension in Hidden States}

In this section, we would like to explore which dimensions of hidden states play a key role when retrieving tools.
During experiments we find a robust model probing (\citealp{model_probe}) method For LLM outputs.
It may help to enhance model interpretability.

As mentioned in Section~\ref{section:method_retrieve}, the Query Encoder $\mathcal{E}_{Q}$ and the Tool Encoder $\mathcal{E}_{T}$ share the same $W_{dim}$ weight ($W_{dim}\in \mathbb{R}^{d}$, with all-ones initialization).
We individually set its learning rate to 0.001, 0.01 and 0.1 for training respectively, and other hyperparameters are kept constant.
The raw $W_{dim}$ weights after training are partially displayed in Figure~\ref{figure:w_tensor_basic}.

\input{figures/w\_tensor\_basic}

It's obvious in Figure~\ref{figure:w_tensor_basic} that parameters change more drastically with larger learning rates.
At a learning rate setting of 0.1, many of the parameters change from the initial 1 to 0.
In addition a similar trend can be vaguely observed for the three folds.

In order to better analyze the commonality of the parameters at different learning rates, we normalize them using Equation~\ref{equation:w_tensor}.
The similar trend of the three folds can be clearly found in Figure~\ref{figure:w_tensor_norm}.
We view dimensions with the common upper value of 3 $W_{dim}$ weights as the key dimensions.
The subscales corresponding to the key dimensions play an important role in tool selection.
\begin{equation}
  \label{equation:w_tensor}
  \hat{W}_{dim} = \frac{W_{dim}-\overline{W}_{dim} }{\sigma_{W_{dim}} } 
\end{equation}

\input{figures/w\_tensor\_norm}

Taking this a step further, we observe the data distribution of raw $W_{dim}$ weights.
The values of 4096 dimensions after the descending order are shown in Figure~\ref{figure:w_tensor_sort}.
As the learning rate increases, the key dimensions become more and more centralized.
At a learning rate of 0.01, 1561 of the 4096 dimensions are weighted more than the initial 1.
We use only these dimensions for tool retrieval.
The TOP1 accuracy has decreased by only 1.4\% from the original 93.8\%, while the TOP5 accuracy remains unchanged.
This provides some evidence that the dimensions of the hidden state of LLM are divided in representing semantic information.

\input{figures/w\_tensor\_sort}

\vspace{-10pt}

%% file: tables/dataset_tool_num.tex
\begin{table}[htb]
\centering
\resizebox{0.35\textwidth}{!}{
\begin{tabular}{lc}
\toprule
\textbf{Dataset}    & \multicolumn{1}{l}{\textbf{Tool Amount}} \\
\midrule
GSM8K-XL           & 4                                     \\
FuncQA              & 13                                    \\
KAMEL               & 234                                   \\
SimpleToolQuestions & 1,836              \\
\bottomrule         
\end{tabular}
}
\caption{Number of tools for several datasets. Dataset SimpleToolQuestions has 999 seen tools and 837 unseen tools (only in the test set).}
\label{table:dataset_tool_amount}
\end{table}

%% file: tables/main_result_math.tex
\begin{table}[htb]
\resizebox{0.48\textwidth}{!}{
\begin{tabular}{lccc}
\toprule
\multicolumn{1}{c}{\multirow{2}{*}{\textbf{Method}}}                  & \multirow{2}{*}{\textbf{GSM8K-XL}} & \multicolumn{2}{c}{\textbf{FuncQA}}    \\
\multicolumn{1}{c}{}                                                    &                                     & \textbf{One-Hop} & \textbf{Multi-Hops} \\
\cmidrule(lr){1-1} \cmidrule(lr){2-2} \cmidrule(lr){3-4}
\textbf{0-shot ChatGPT}                                                 & 0.17                                & 0.55             & 0.09                \\
\midrule
\textbf{0-shot Prompting} \emph{LLaMA}                                   & 0.04                                & 0.05             & 0.00                \\
\textbf{CoT Prompting} \emph{LLaMA}                                   & 0.00                                & 0.00             & 0.00                \\
\textbf{ToolkenGPT} \emph{LLaMA}                                   & 0.18                                & 0.48             & 0.06                \\
\textbf{CoTools(ours)} \emph{LLaMA}                                    & \textbf{0.19}                                & \textbf{0.53}             & \textbf{0.07}         \\
\midrule
\textbf{0-shot Prompting} \emph{Mistral}                                   & 0.14                                & 0.17             & 0.04                \\
\textbf{CoT Prompting} \emph{Mistral}                                   & 0.10                                & 0.20             & 0.06                \\
\textbf{CoTools(ours)} \emph{Mistral}                                    & \textbf{0.42}                                & \textbf{0.63}             & \textbf{0.07}         \\ \bottomrule         
\end{tabular}
}
\caption{Main results on Numerical Reasoning Benchmarks. Round Acc metric (round float numbers to two decimals) for GSM8K-XL and FuncQA One-Hop. Approx Acc metric (allow 0.1\% error) for FuncQA Multi-Hops. \emph{LLaMA} refers to \emph{LLaMA2-7B-Chat}. \emph{Mistral} refers to \emph{Mistral-7B-Instruct-v0.2}. Every result is from a single run.}
\label{table:main_result_math}
\end{table}

%% file: tables/main_result_KBQA.tex
\begin{table}[htb]
\resizebox{0.48\textwidth}{!}{
\begin{tabular}{lcccc}
\toprule
\multicolumn{1}{c}{\multirow{2}{*}{\textbf{Method (\%)}}} & \multicolumn{2}{c}{\textbf{KAMEL}} & \multicolumn{2}{c}{\textbf{STQuestions}} \\
\multicolumn{1}{c}{}                                   & \textbf{SUP}     & \textbf{SYN}    & \textbf{Seen}          & \textbf{Unseen}         \\
\cmidrule(lr){1-1} \cmidrule(lr){2-3} \cmidrule(lr){4-5}
\textbf{ToolkenGPT} \emph{LLaMA}                                   & 93.4             & 20.6            & 23.8                   & 0.0                     \\
\textbf{CoTools(Ours)} \emph{LLaMA}                                    & 93.8             & 43.6            & 35.1                   & 10.4                    \\
\bottomrule         
\end{tabular}
}
\caption{Main results on KBQA Benchmarks. The eval metric is the accuracy of tool selection. Because of the large number of tools, we do not evaluate the Prompting method which is inefficient and hard to guarantee results. \emph{LLaMA} refers to \emph{LLaMA2-7B-Chat}.}
\label{table:main_result_KBQA}
\end{table}

%% file: figures/data_synthesis_sup.tex
\begin{figure}[htb]    
	\centering   
	\includegraphics[width=0.98\linewidth]{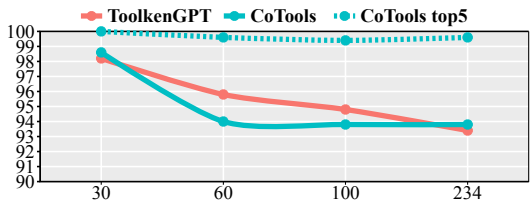}    
	\caption{Performances of the methods with foundation model fine-tuned by KAMEL(\emph{sup}). The X-axis represents the number of tools in the tool pool. The Y-axis represents the tool selection accuracy.} 
	\label{figure:data_synthesis_sup}
\end{figure}

%% file: figures/data_synthesis_syn.tex
\begin{figure}[htb]    
	\centering   
	\includegraphics[width=0.98\linewidth]{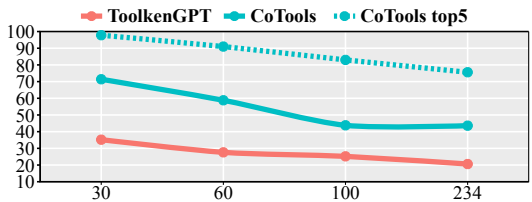}    
	\caption{Performances of the methods with foundation model fine-tuned by KAMEL(\emph{syn}). The rest of the settings are the same as in Figure~\ref{figure:data_synthesis_sup}.} 
	\label{figure:data_synthesis_syn}
\end{figure}

%% file: figures/scaling_law.tex
\begin{figure}[htb]    
	\centering   
	\includegraphics[width=0.98\linewidth]{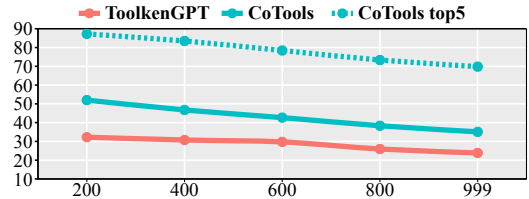}    
	\caption{Tool selection result with massive tools. The X-axis represents the number of tools in the tool pool. The Y-axis represents the tool selection accuracy.}
	\label{figure:scaling_law}
\end{figure}

%% file: figures/unseen_error_analysis.tex
\begin{figure}[htb]    
	\centering   
	\includegraphics[width=0.98\linewidth]{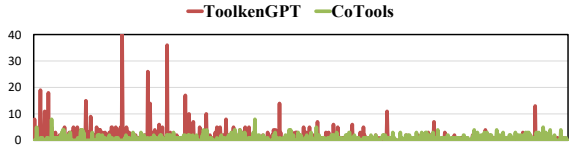}    
	\caption{Error analysis for instances with unseen tools. We count the number of instances where the unseen tool should have been called but the seen tool is called instead. The X-axis represents the 999 seen tools. The Y-axis represents the number of times the seen tool is incorrectly called. (It is recommended to view the color-printed version.)}
	\label{figure:unseen_error_analysis}
\end{figure}

%% file: figures/w_tensor_basic.tex
\begin{figure}[htb]    
	\centering   
	\includegraphics[width=0.98\linewidth]{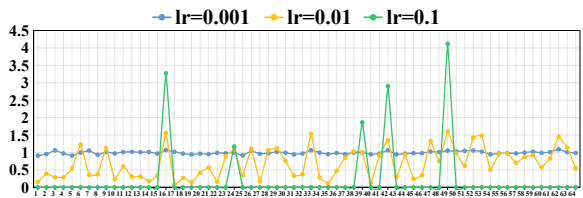}    
	\caption{Raw $W_{dim}$ weight segment (only the first 64 dimensions). The weights are initialized with all ones.} 
	\label{figure:w_tensor_basic}
\end{figure}

%% file: figures/w_tensor_norm.tex
\begin{figure}[htb]    
	\centering   
	\includegraphics[width=0.98\linewidth]{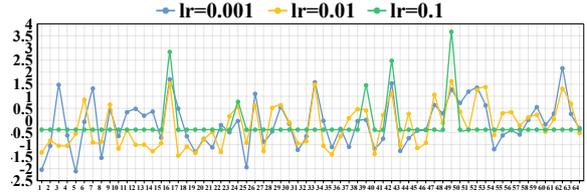}    
	\caption{Normalized $W_{dim}$ weight segment (only the first 64 dimensions) by equation~\ref{equation:w_tensor}.}
	\label{figure:w_tensor_norm}
\end{figure}

%% file: figures/w_tensor_sort.tex
\begin{figure}[htb]    
	\centering   
	\includegraphics[width=0.92\linewidth]{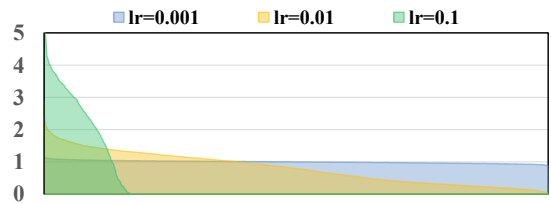}    
	\caption{Sorted raw $W_{dim}$ weight. It's trained with the dataset KAMEL.}
	\label{figure:w_tensor_sort}
\end{figure}

%% file: sections/conclusion.tex
\section{Conclusion}
\label{sec:conclusion}

In this paper, we present a novel CoT tool learning method CoTools which is based on frozen LLMs.
CoTools incorporates tool invocation into the process of generating answers to the model while maintaining the original generalized ability of the foundation model.
It leverages the powerful semantic representation capability of the foundation model to determine whether to call a tool and what tool to call.
Experiments on Benchmarks of numerical reasoning and KBQA tasks show that our method performs well in different scenarios.
In particular, its tool selection capability is much improved compared to the baseline.
In addition, we roughly explore the role that different hidden states dimensions of model output play in the tool selection process, which helps to enhance model interpretability.
We believe that our ideal Tool Learning agent framework based on frozen LLMs with its practical realization method CoTools can be useful in real-world applications and even drive further development of Tool Learning.

%% file: sections/limitations.tex
\section*{Limitations}
\label{sec:limitations}

Research on LLM Tool Learning is still on its early stage.
Although the open source community already has a number of Tool Learning datasets, a large proportion of them are too simple or of low quality. 
As a whole, our research is thus limited.

One limitation is that we do not conduct experiments for tools containing multiple return values.
In this situation, we need to select the return value that should be added to the answer.
Our solution is to train a similar Return Value Encoder $\mathcal{E_{R}}$ like $\mathcal{E_{Q}}$ in Section~\ref{section:method_retrieve}.
Unfortunately it cannot not be evaluated at the moment because of the lack of relevant CoT Tool Learning datasets with this kind of tools.
For the selection of return values for this type of tools, a temporary alternative to the prompting scheme is currently available.

Another possible limitation is that we do not attempt the complete Tool Learning process on a large-scale real-world toolset.
A Tool Learning dataset with large-scale realizable tools still does not exist.
The dataset ToolBench (\citealp{ToolLLM}) comes closest to this goal.
However the format of its gold data is too cumbersome and not well filtered.

%% file: sections/appendix.tex
\appendix

\section{Hyper Parameter}
\label{sec:appendix_hyper_parameter}

Hyper parameters for fine-tuning is shown in Table~\ref{table:hyper_parameter}.
When fine-tuning the Tool Retriever with GSM8K-XL, we adjust the batch size because of the CUDA Out Of Memory Error reported.
Back propagation frequency in fine-tuning Tool Retriever with FuncQA is also adapted since it only has 611 pieces of data.

\input{tables/hyper_parameter}

For learning rate of $W_{dim}$, we generally set it to 0.01.
In fact, $W_{dim}$ has little effect on the fine-tuning effect and can be removed by adjusting "tensor\_weighting" to "false" in the settings of our source code.
It is mainly used to explore the semantic information contained in hidden states.

\section{Dataset}

\subsection{Dataset Split}

\input{tables/dataset_detail}

The specific details of the 4 datasets are shown in Table~\ref{table:dataset_detail}.
Further additional explanation of the details of the dataset is provided below:
The test set of dataset FuncQA contains 60 single-hop problems and 68 multi-hop problems.
The training set of KAMEL contains 19,000 artificially constructed gold standard data (sup) and 8,095 ChatGPT synthesized data (syn).
They are separated for training.
The dataset STQuestions contains 999 seen tools and 837 unseen tools.
Its test set is also divided into 1,707 questions for seen tools and 1,066 questions for unseen tools.

\subsection{Dataset License}

\noindent\textbf{GSM8K-XL}: MIT License.

\noindent\textbf{FuncQA}: Apache License 2.0.

\noindent\textbf{KAMEL}: The Creative Commons Attribution-Noncommercial 4.0 International License.

\noindent\textbf{STQuestions}: Creative Commons Attribution 3.0 License.

\section{Prompt}

To ensure experimental fairness, we all use the prompt provided in the dataset or baseline for inference.
The other Prompts utilized are shown below.

Tool Prompt : 
\begin{tcolorbox}[boxrule=0pt]
tool name: [\emph{Tool Name}], tool description: [\emph{Tool Description}]
\end{tcolorbox}

Retrieval Prompt of the dataset GSM8K-XL:
\begin{tcolorbox}[boxrule=0pt]
\text{[\emph{Query}]} Let's think step by step.[\emph{Answer Fragment}]
\end{tcolorbox}

Retrieval Prompt of the dataset FuncQA:
\begin{tcolorbox}[boxrule=0pt]
Q: [\emph{Query}]

A: [\emph{Answer Fragment}]
\end{tcolorbox}

Retrieval Prompt of the dataset KAMEL and STQuestions:
\begin{tcolorbox}[boxrule=0pt]
Question: [\emph{Question}]

Answer: The answer is
\end{tcolorbox}

\section{Computational Resource}

For LLaMA2-7B-Chat, the computing resources we use are 2 $\times$ NVIDIA V100 32G.
Training the Tool Retriever with the GSM8K-XL dataset (5,054 cases) takes approximately 2 hours per epoch.

\section{Safeguarding Statement}

In this paper, we focus on the application of Tool Learning to solve numerical reasoning and knowledge retrieval tasks.
We do not believe it poses any political, ethical or legal risk.
In the future, we will also explore how LLMs can be better integrated with Tool Learning to serve human society.

%% file: tables/hyper_parameter.tex
\begin{table}[htb]
\centering
\resizebox{0.48\textwidth}{!}{
\begin{tabular}{lcccc}
\toprule
\textbf{Module}              & \textbf{Epoch} & \textbf{Learning Rate} & \textbf{Batch Size} & \textbf{Accumulation Step} \\
\midrule
\textbf{Judge}               & 3              & 1e-5                   & 8                   & 16                         \\
\textbf{Retriever}           & 10             & 1e-4                   & 16                  & 12                         \\
\textbf{Retriever(GSM8K-XL)} & 10             & 1e-4                   & 12                  & 16        \\
\textbf{Retriever(FuncQA)}   & 10             & 1e-4                   & 8                   & 6  \\
\bottomrule         
\end{tabular}
}
\caption{Hyper parameters used in fine-tuning with foundation models LLaMA2-7B-Chat and Mistral-7B-Instruct-v0.2.}
\label{table:hyper_parameter}
\end{table}

%% file: tables/dataset_detail.tex
\begin{table}[htb]
\centering
\resizebox{0.35\textwidth}{!}{
\begin{tabular}{lrrrr}
\toprule
\textbf{Dataset}     & \multicolumn{1}{c}{\textbf{Tool}}                 & \multicolumn{1}{c}{\textbf{Train}}                     & \multicolumn{1}{c}{\textbf{Dev}} & \multicolumn{1}{c}{\textbf{Test}}                     \\
\midrule
\textbf{GSM8K-XL}    & 4                                                 & 5,054                                                  & 1,000                            & 568                                                   \\
\midrule
\textbf{FuncQA}      & 13                                                & 611                                                    & 39                               & \begin{tabular}[c]{@{}r@{}}60\\ 68\end{tabular}       \\
\midrule
\textbf{KAMEL}       & 234                                               & \begin{tabular}[c]{@{}r@{}}19,000\\ 8,095\end{tabular} & 1,000                             & 500                                                   \\
\midrule
\textbf{STQuestions} & \begin{tabular}[c]{@{}r@{}}999\\ 837\end{tabular} & 10,483                                                 & 1,707                             & \begin{tabular}[c]{@{}r@{}}1,707\\ 1,066\end{tabular} \\
\bottomrule         
\end{tabular}
}
\caption{Detailed information of 4 datasets, including the number of tools for each dataset and the distribution of data across the training, development, and test sets.}
\label{table:dataset_detail}
\end{table}